\documentclass[letterpaper,journal]{IEEEtran}
\usepackage[utf8]{inputenc}
\usepackage{amsmath}
\usepackage{amssymb}
\usepackage{cite}
\usepackage{graphicx}
\usepackage{booktabs}
\usepackage{array}
\usepackage{multirow}
\usepackage[table]{xcolor}
\usepackage{capt-of}
\usepackage{etoolbox}
\usepackage{hyperref}
\usepackage{placeins}

\makeatletter
\patchcmd{\@makecaption}
  {\normalfont\footnotesize \scshape #2}
  {\normalfont\footnotesize #2}
  {}{}
\makeatother

\newcommand{\topic}[1]{%
    \par\noindent\textbf{#1} %
}

\title{\LARGE\bfseries
ViTacPhys: Physical Property-Aware Grasping from Human Visual-Tactile Demonstrations
}

\author{Yiwen Liu, Yujun Zhu, Kui Jia, Zhao Liao, Yangwei You, and Shuaijun Wang}

\begin{document}

\twocolumn[{%
\renewcommand\twocolumn[1][]{#1}%
\maketitle

\begin{center}
    \vspace{-2.0em}
    \includegraphics[width=0.98\textwidth]{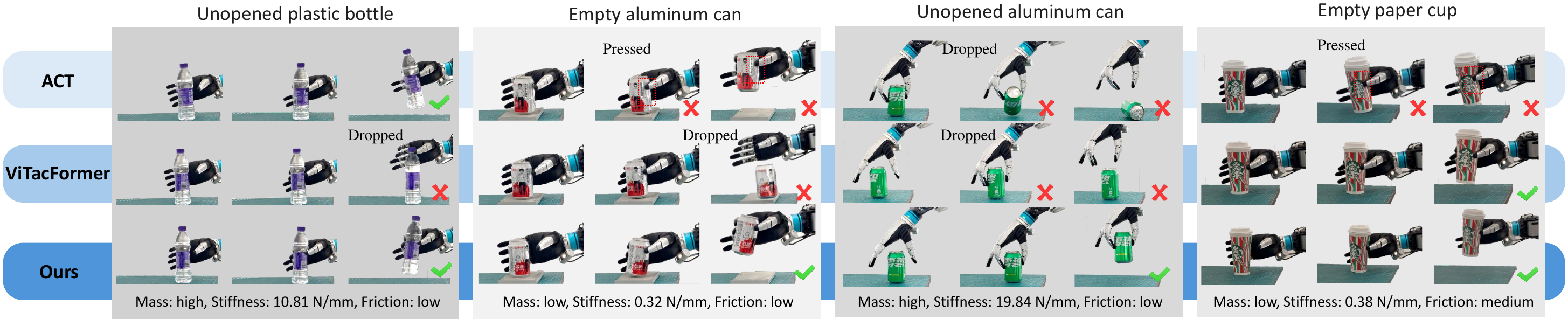}
    \vspace{-0.6em}
    \captionof{figure}{
    ViTacPhys predicts object mass class, stiffness, and friction-coefficient class from visual--tactile observations and conditions a downstream adaptive grasping policy on these estimates. The examples compare grasping outcomes for visually similar objects with different physical properties against ACT and ViTacFormer~\cite{act,2025vitacformer}.
    }
    \label{fig:teaser}
    \vspace{-0.4em}
\end{center}%
}]
\begingroup
\renewcommand{\thefootnote}{}
\footnotetext{All authors are with Xiaomi Robotics. Corresponding author: Shuaijun Wang (e-mail: wukongwoong@gmail.com).}
\endgroup

\begin{abstract}
Recent vision-based action models show strong capabilities in complex manipulation, but they rarely use explicit object physical properties to adapt manipulation policies. We introduce \textbf{ViTacPhys}, a visual--tactile framework and data acquisition system for predicting object mass, stiffness, and friction-coefficient class from human manipulation demonstrations. Trained on data from $60$ rigid and deformable objects, ViTacPhys combines temporal visual--tactile modeling, cross-attention-based multimodal fusion, and a VLM-derived semantic prior. On seen objects, it achieves $97.2\%$ mass accuracy, $98.8\%$ friction-coefficient accuracy, and $5.51\%$ stiffness MAPE. On held-out objects from known categories, it achieves $87.5\%$ mass accuracy, $97.5\%$ friction-coefficient accuracy, and $9.08\%$ stiffness MAPE. We transfer ViTacPhys from the human to the robot domain using limited teleoperation data, robot-style video augmentation, and matched-action human demonstrations, and deploy it as an online module for adaptive grasping. The resulting physical-property-conditioned policy achieves $95.0\%$ total grasping success on ID objects and $83.4\%$ on OOD objects. On the common OOD objects successfully grasped by both methods, its force profile is more consistent with human teleoperation than that of ACT. These results provide a system-level feasibility study of explicit physical-property estimates for real-world adaptive grasping. Project page: \href{https://vitacphys.github.io/ViTacPhys/}{https://vitacphys.github.io/ViTacPhys/.}

\end{abstract}


\begin{table*}[!t]
\caption{Comparison of physical-property datasets.}
\label{tab:dataset_comparison}
\centering
\scriptsize
\renewcommand{\arraystretch}{1.0}
\setlength{\tabcolsep}{2pt}
\begin{minipage}[t]{0.485\textwidth}
\centering
\textbf{Implicit Labels}
\resizebox{\linewidth}{!}{%
\begin{tabular}{>{\centering\arraybackslash}p{2.0cm}
                >{\centering\arraybackslash}p{0.9cm}
                >{\centering\arraybackslash}p{6.8cm}}
\toprule
\textbf{Dataset} & \textbf{\#Obj.} & \textbf{Properties / Modalities} \\
\midrule
Physics 101~\cite{phys101}
& 101
& Mass, density, elasticity // RGB-D video \\
OmniPush~\cite{omnipush}
& 250
& Mass, friction // RGB-D, motion \\
SPS~\cite{synesthesia}
& 400+
& Material, friction, compliance // RGB, tactile signals \\
Fabrics~\cite{yuan2017connecting}
& 118
& Material, stiffness, thickness // RGB-D, GelSight \\
\bottomrule
\end{tabular}}
\end{minipage}
\hfill
\begin{minipage}[t]{0.485\textwidth}
\centering
\textbf{Explicit Labels}
\resizebox{\linewidth}{!}{%
\begin{tabular}{>{\centering\arraybackslash}p{2.0cm}
                >{\centering\arraybackslash}p{0.9cm}
                >{\centering\arraybackslash}p{6.8cm}}
\toprule
\textbf{Dataset} & \textbf{\#Obj.} & \textbf{Properties / Modalities} \\
\midrule
Image2Mass~\cite{image2mass}
& 150K/56
& Volume, density, mass // RGB, bounding box \\
Tactile glove~\cite{sundaram2019learning}
& 26
& Object shape, mass // whole-hand tactile images \\
Octopi~\cite{robotic_perception,yu2024octopi}
& 35
& Hardness, elasticity, roughness // RGB, GelSight images, language \\
\textbf{ViTacPhys (Ours)}
& \textbf{60}
& \textbf{Mass, stiffness, friction coefficient // wrist RGB, tactile maps} \\
\bottomrule
\end{tabular}}
\end{minipage}
\vspace{-0.8em}
\end{table*}

\vspace{-0.4em}
\section{Introduction}
\label{sec:intro}

Recent Vision-Language-Action (VLA) models~\cite{pi_05} and World Action Models (WAMs)~\cite{dreamzero} have shown impressive performance in manipulation tasks. However, manipulating diverse everyday objects in real-world environments, such as logistics robots handling packages or household robots organizing tabletops, requires not only knowledge about object categories, geometry, and poses~\cite{wang2022learning}, but also the ability to perceive and adapt to physical properties such as mass,  stiffness, and friction coefficient.
For example, mass~\cite{motion_kinematics,image2mass,depth2mass} and friction coefficient~\cite{force,wang2020swingbot,sundaralingam2021hand,xu2019densephysnet} determine required grasping tangential force by preventing the object from slipping, while  stiffness determines the appropriate normal force to avoid undesired object deformation.~\cite{phys101,robotic_perception,kruzliak2024interactive}. When manipulating a delicate paper cup, excessive force may deform or crush the paper cup, whereas insufficient force may cause it to slip and fall~\cite{sugaiwa2010methodology}. Humans have rich prior knowledge that enables a form of visuo-tactile synesthesia: even from visual observation alone, they can often predict how an object will feel, infer its physical properties, and formulate appropriate manipulation strategies~\cite{bmw2025predictive}. Even for unseen objects, humans can rapidly refine these estimates and adapt actions by integrating visual cues with dense tactile feedback obtained through brief contact or hefting motions~\cite{twostream,robotic_perception,kruzliak2024interactive,wang2025tacrefinenet}. In contrast, enabling robots to acquire such synesthesia-like cross-modal perceptual capabilities and perform physical-property-aware manipulation remains a central challenge~\cite{synesthesia,yuan2017connecting}.

Prior works learn object physical properties either by acquiring human-like synesthesia priors from large-scale visual and tactile data~\cite{image2mass,sundaram2019learning,synesthesia,yuan2017connecting}, for example through contrastive representation learning, or by designing interactive actions such as pushing~\cite{xu2019densephysnet}\cite{bmw2023push}\cite{bmw2025predictive}\cite{omnipush}, pulling~\cite{bmw2025predictive}\cite{force}, sliding~\cite{sim2023object} \cite{xu2019densephysnet}, poking~\cite{chuo2016learning}. Although these robot-specific interactions provide useful physical evidence, they are often decoupled from the target manipulation task and difficult to deploy in real-world settings. Human data offers a scalable and cost-effective way to collect rich manipulation experience~\cite{egoscale,egomimic}. Humans can provide precise manipulation demonstrations for objects with diverse physical properties, thereby producing accurate visual and tactile observations as inputs for physical property prediction. However, only a few works have explored how to leverage human data for physical property prediction~\cite{sundaram2019learning,force,motion_kinematics}. More importantly, most existing physical property prediction methods are treated as offline estimators, rather than being directly transferred into downstream adaptive grasping policies.

We therefore predict mass, stiffness, and friction-coefficient from temporally structured visual-tactile evidence. Vision supplies appearance, geometry, material, and fill-state cues; tactile maps capture local mechanics; and temporal content and flow describe the evolving interaction. ViTacPhys combines this model with a synchronized dataset of $1{,}800$ human grasping demonstrations from $60$ rigid and deformable objects. A semantic prior generated from five strictly pre-contact RGB frames complements the post-contact sequence without exposing the VLM to contact deformation or motion. For human-to-robot transfer, we combine limited teleoperation with matched-action human data and robot-style RGB augmentation~\cite{egoscale}. The predicted properties condition an ACT-style policy~\cite{act}, allowing us to evaluate how this conditioning is associated with grasp success, force adaptation, and generalization to visually similar objects with different properties.
The main contributions of this paper are summarized as follows:
\begin{itemize}
    \item We introduce a synchronized human visual--tactile dataset containing $60$ objects, together with measurement protocols for mass, stiffness, and silicone-contact friction coefficient.
    \item We propose a temporal multimodal predictor that combines visual and tactile content, interaction flow, and a pre-contact VLM semantic prior. It achieves $87.5\%$ mass accuracy, $97.5\%$ friction-coefficient accuracy, and $9.08\%$ stiffness MAPE on held-out objects from known categories.
    \item We integrate the predictor into a complete dexterous-robot grasping system as a feasibility study. In our evaluation, the conditioned policy is associated with clean-success rates $12.5$ percentage points higher than ACT on ID objects and $38.9$ percentage points higher on OOD objects.
\end{itemize}

\vspace{-0.6em}
\section{Related Works}
\label{sec:relatedwork}

\subsection{Object Physical Properties Datasets}
Physical-property datasets can be broadly divided into two categories: datasets with explicit labels and datasets with implicit labels. Representative datasets are summarized in Table~\ref{tab:dataset_comparison}. Although implicit datasets~\cite{phys101, omnipush, synesthesia, yuan2017connecting} are generally larger in scale because they avoid explicit physical-property labeling, they are often constrained by specific experimental scenarios. For example, Physics 101 collects data through predefined interactions, such as ramp experiments in which objects slide down an inclined plane~\cite{phys101}. In contrast, explicit datasets require carefully designed protocols and instruments to obtain ground-truth physical properties; therefore, they inevitably involve measurement noise and are typically smaller in scale~\cite{image2mass,sundaram2019learning,robotic_perception,yu2024octopi}.

Existing datasets cover diverse sensing modalities, some of which can be combined to analytically derive physical properties, such as estimating stiffness from position and force measurements~\cite{sim2023object,yang2025learning}. 
Although force measurements are often treated as tactile modality~\cite{synesthesia,twostream,bmw2025predictive,sim2023object}, only a few works directly use tactile images as model inputs~\cite{robotic_perception}. In addition, many datasets rely on high-precision vision-based tactile sensors such as GelSight~\cite{yuan2017connecting,robotic_perception,wang2020swingbot,yang2025learning}, which are difficult to mount on dexterous hands. As shown in Table~\ref{tab:dataset_comparison}, mass-related, stiffness-related, and friction-related properties are among the most manipulation-relevant physical attributes. In this work, we focus on predicting mass, stiffness, and static friction coefficient.

\vspace{-0.6em}
\subsection{Learning Object Physical Properties}  

Explicit datasets enable supervised learning of physical properties. Purely visual methods can provide global priors before contact, for example by estimating mass from visual appearance~\cite{image2mass,depth2mass}. However, these methods can fail on disguised materials or unseen objects, while tactile provides complementary information beyond visual appearance. Existing methods have used fingertip tactile images~\cite{twostream}, whole-hand tactile sensing~\cite{sundaram2019learning}, or GelSight-based tactile sensing~\cite{yang2025learning}, together with ResNet-like networks~\cite{resnet}, factor graphs~\cite{sundaralingam2021hand}, or dual Kalman filters~\cite{sim2023object}, to predict object properties such as mass, shape, stiffness, and friction coefficients.
Although tactile-only methods provide rich contact feedback, they require physical contact and are limited to local sensing. Multimodal fusion methods predict physical properties from visual, tactile, and force signals during pushing~\cite{bmw2023push,bmw2025predictive}, pulling~\cite{force}, or squeezing~\cite{kruzliak2024interactive} interactions using dual-channel filters or Bayesian networks. Recent works further leverage the reasoning ability and prior knowledge of VLMs to assist physical property prediction~\cite{robotic_perception,liu2025does,gaussian,guo2025phygrasp}.

Implicit datasets can provide stronger priors through self-supervised learning. Contrastive learning~\cite{yuan2017connecting} and adversarial learning~\cite{synesthesia} have been used to align visual images with GelSight tactile images or force signals, enabling a human-synesthesia-like capability. Another line of work constructs self-supervised objectives from implicit signals related to physical properties. For example, some methods use human poses from optical motion capture~\cite{force}, robot proprioception~\cite{simreal2025learning}, or videos from predefined experiments such as ramps and springs~\cite{phys101}, and optimize physical property in simulation to reproduce real object motion. However, these methods remain affected by the sim-to-real gap. Other works learn implicit physical representations through self-supervised tasks based on poking, sliding, collision, random pushing, or controlled throwing interactions, such as predicting post-poking images~\cite{chuo2016learning}, optical flow~\cite{xu2019densephysnet}, or throwing angles~\cite{omnipush}. Most existing studies remain limited to offline classification, parameter estimation, or predefined interaction tasks. 

Moreover, human demonstrations contain subtle but consistent action patterns that reflect object physical properties, and their similarity to dexterous-hand manipulation makes them promising for transferring such knowledge to robots. However, only a few works have explored learning physical properties from human data~\cite{sundaram2019learning,force,motion_kinematics}. To address this gap, we learn physical properties from human grasping demonstrations using visual and tactile observations, transfer the learned knowledge to the robot domain, and integrate the predicted properties into downstream adaptive grasping policies.

\vspace{-1.0em}
\subsection{Adaptive Grasping}
Classical impedance-control-based methods can adapt grasping behavior based on contact feedback, but they often require carefully designed controllers and high-resolution tactile or force sensing~\cite{wimboeck2006passivity, impedence}, which limits their generality across diverse objects and tasks. In contrast, imitation learning with tactile or force inputs provides a more general framework for fine-grained manipulation. Recent works have incorporated tactile images through MoE architectures~\cite{2025omnivtla} or contrastive learning with other modalities~\cite{2505forcevla}, and have explored multimodal fusion mechanisms such as self-attention~\cite{2025softgrasp} and cross-attention~\cite{2025fbi}. Within the ACT framework~\cite{act}, FTACT~\cite{FTACT} directly introduces raw force signals and shows that force variations can indicate different execution stages. ViTacFormer~\cite{2025vitacformer} further improves visual-tactile policy learning by predicting the next tactile frame. SoftGrasp~\cite{2025softgrasp} is closely related to our application, as it studies adaptive grasping for objects with different softness. In contrast, our work explicitly predicts grasp-relevant physical properties from human visual-tactile demonstrations and uses them as structured priors for downstream adaptive grasping.

\section{Human Visual--Tactile Physical-Property Dataset}
\label{sec:dataset}

\begin{figure}[!t]
\centering
\includegraphics[width=.94\linewidth]{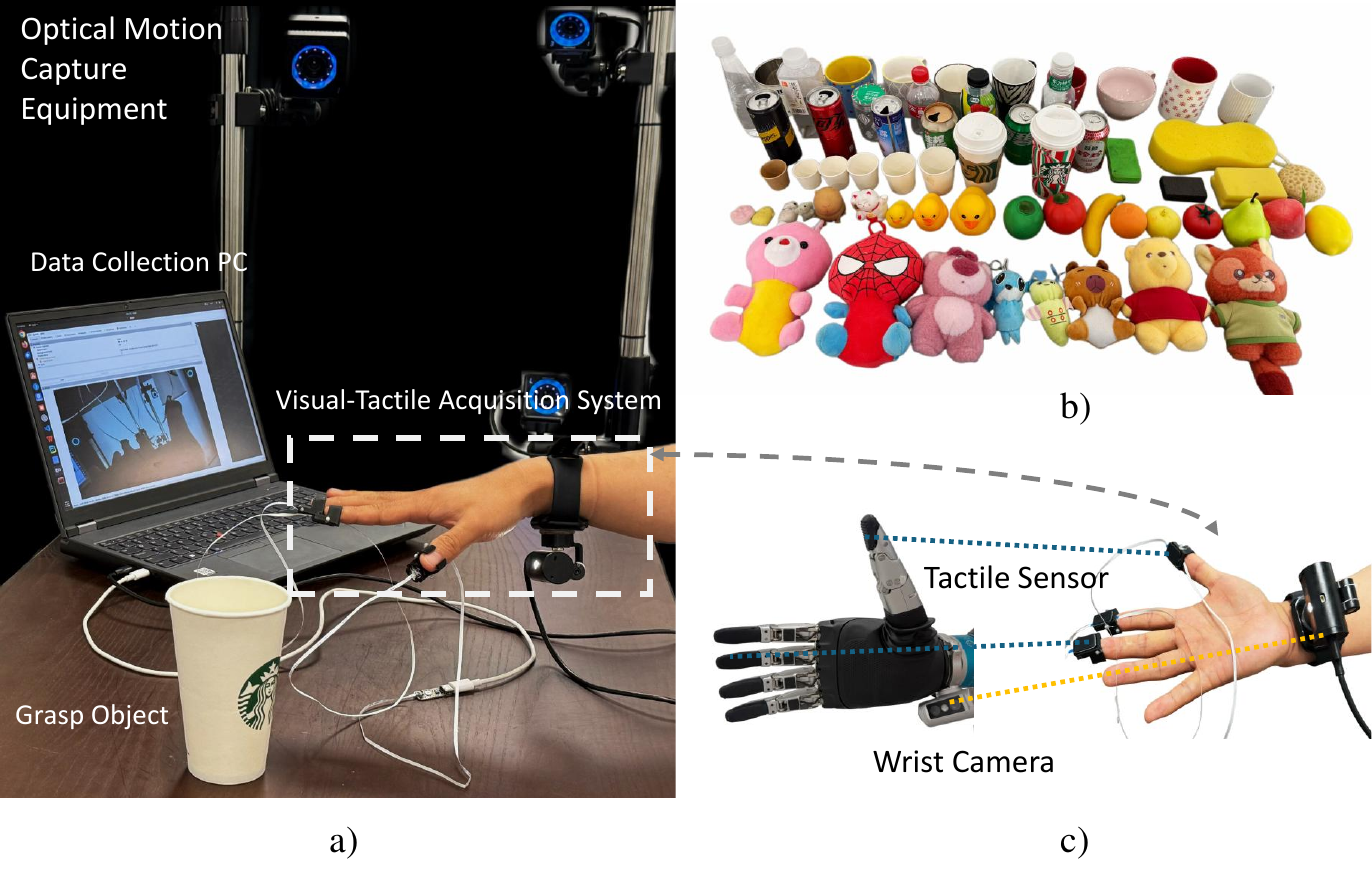}
\vspace{-0.6em}
\caption{Panel (a) shows the human visual--tactile data collection setup, which synchronizes wrist-view video, fingertip pressure maps, and motion-capture measurements. Panel (b) shows the everyday objects in the dataset, ranging from rigid mugs to soft rubber toys. Panel (c) shows the corresponding sensor locations on the wearable device and robot platform.
}
\vspace{-0.3em}
\label{figs:data_collect}
\end{figure}


\begin{figure}[!t]
\centering
\includegraphics[width=.98\linewidth]{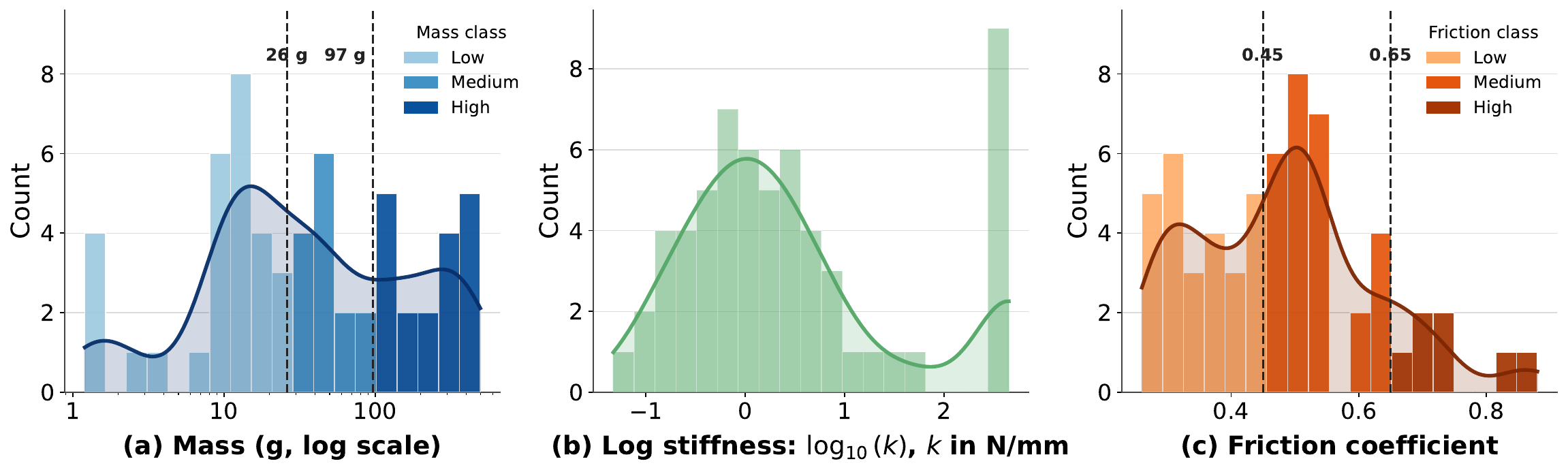}
\vspace{-0.6em}
\caption{Object-level label distributions in the ViTacPhys dataset. Mass and friction coefficient are discretized into three ordered classes, whereas stiffness is a continuous regression target shown on a $\log_{10}$ scale. The displayed dataset-level class boundaries are fitted once using all annotated objects and fixed across evaluation splits.}
\vspace{-0.3em}
\label{figs:data_distribution}
\end{figure}

\begin{figure*}[!t]
\centering
\includegraphics[width=.99\linewidth]{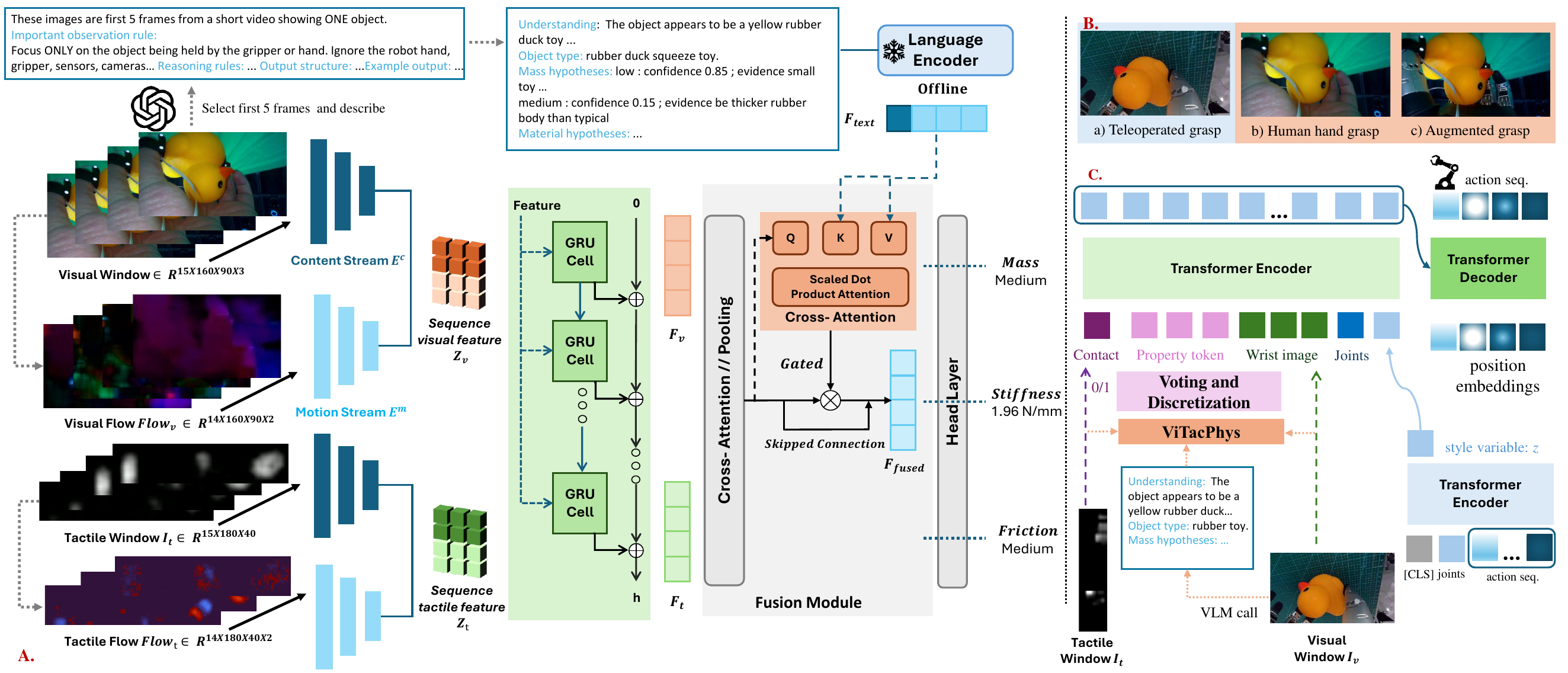}
\vspace{-0.6em}
\caption{Overview of ViTacPhys. (A) The predictor combines temporal visual--tactile observations, flow features, and a VLM-derived semantic prior computed from pre-contact RGB frames. (B) Human-to-robot transfer uses robot teleoperation, matched-action human demonstrations, and visually augmented human demonstrations. (C) During deployment, the VLM prior is computed before contact. After contact, the system immediately begins rolling prediction from a 30-frame visual--tactile queue; unavailable initial entries repeat the earliest post-contact observation before 15 frames and 14 adjacent flow fields are sampled. Cumulative temporal voting stabilizes the predicted classes and stiffness bin before they are passed to the downstream policy.}

\label{figs:model1}
\end{figure*}
We build an in-house human visual--tactile dataset for learning physical properties from natural grasping and describe its acquisition, annotation, interactions, and composition.
\vspace{-0.3em}
\subsection{Data Collection System}
\label{sec:dataset_system}
The wearable system in Fig.~\ref{figs:data_collect}(a) mounts pressure arrays on the thumb, index, and middle fingertips, an RGB camera on the wrist, and motion-capture markers near the fingertips for displacement measurement. Each fingertip sensor produces a pressure image according to its tactile taxel layout. Images from the three fingers are concatenated and post-processed into the tactile input $I_{\mathrm{t}}\in\mathbb{R}^{180\times40}$. The wrist-mounted RGB stream is post-processed into the visual input $I_{\mathrm{v}}\in\mathbb{R}^{160\times90\times3}$. To increase data diversity, we only require a coarse visual-scale alignment between the glove and the wrist-mounted camera during setup.

\vspace{-0.6em}
\subsection{Data Annotation}
We use separate protocols to annotate the three physical properties.

\topic{Mass.} Mass is measured five times with a $0.1$~g-resolution scale; the mean is the label and the sample standard deviation records uncertainty.
\topic{Stiffness.} During quasi-static pre-yield pinching, the thumb and index finger move toward each other while their motion is kept as normal to the object surface as possible. Motion capture measures the change in relative thumb--index separation along the contact-normal direction, and calibrated tactile readings provide the normal force from each finger after fourth-order Butterworth filtering. Fitting uses the stable loading interval to exclude contact transients and unloading hysteresis.

For each fingertip, $S$ is the mean active-taxel 8-bit reading divided by $255$. Known loads fit $F=aS+b$ in newtons. For the stiffness measurement, $F$ is the sum of the calibrated normal forces from the thumb and index fingers, and $x$ is the corresponding relative-finger displacement. For normalized-force comparisons, the grasp signal is the sum of the normalized readings $S$ from the thumb, index, and middle fingers; force plots in newtons instead sum the corresponding calibrated forces $F$. For displacement--force pairs $\mathbf{x}$ and $\mathbf{F}$, stiffness is
\begin{equation}
k=\frac{\sum_i(x_i-\bar{x})(F_i-\bar{F})}{\sum_i(x_i-\bar{x})^2}.
\end{equation}
We report $k$ in N/mm as the mean of five trials and retain the inter-trial standard deviation. Because it includes object deformation and hand, sensor-mount, and contact compliance, this operational stiffness is not an intrinsic material constant.

\topic{Friction Coefficient.}
The object is placed on a silicone-coated inclined plane matching the tactile contact material. At the critical sliding angle $\theta$, the contact-pair coefficient is $\mu_s=\tan(\theta)$; it is not an object-only property. Five repetitions yield the object-level mean and sample standard deviation, avoiding the use of category or appearance as friction ground truth.

\vspace{-0.6em}
\subsection{Dataset Overview}

The dataset contains $60$ everyday objects in seven categories with varied shape, material, surface, filling, and rigidity (Fig.~\ref{figs:data_collect}(b)). Each has mass, stiffness, and silicone-contact friction-coefficient labels. 

All properties are measured continuously, but the arrays capture normal pressure rather than tangential force, weakly constraining exact mass and friction-coefficient regression. Because these properties guide grasping mainly through coarse ranges, they are ordered low-medium-high targets; stiffness remains continuous. Before any protocol- or seed-specific split is formed, one-dimensional $k$-means~\cite{kmeans} is fitted once to the annotated mass or friction-coefficient values of all benchmark objects. This establishes a pre-defined dataset-level label ontology rather than a boundary estimated from a particular test set. The ontology is never re-estimated or selected using test performance, while the test files themselves vary across protocols and independent seeds. Ordered centroids define the low, medium, and high classes, and their midpoints are then held fixed for every training, validation, and test split. Figure~\ref{figs:data_distribution} shows these fixed boundaries.

To enrich the physical information captured during interaction, the ViTacPhys dataset uses two collection protocols. In the vertical grasping protocol, the object is pinched from the table and lifted vertically at an approximately constant speed, producing stable tactile signals after contact. In the shaking-grasp protocol, the participant grasps the object and swings it laterally with a large amplitude, introducing dynamic interaction cues and varying the relationship between gravity and the fingertip normal direction, which can facilitate mass estimation.

Compared with teleoperation data, human demonstrations can capture natural action adaptations to different physical property combinations, enable efficient large-scale data collection, and easily support the two interaction protocols described above.
For each object, we collect $15$ trials per protocol with different approach directions and hand postures, resulting in $1800$ human grasping demonstrations collected by one participant. To encourage early physical property inference, each synchronized visual-tactile sequence is processed into a $1$-second clip at $30$~Hz.

\vspace{-0.6em}
\section{Method}
\label{sec:method}

Figure~\ref{figs:model1} presents ViTacPhys and its integration into downstream adaptive grasping. Secs.~\ref{sec:vt-encoders}--\ref{sec:training-loss} describe the predictor, Sec.~\ref{sec:human-to-robot} describes transfer to the robot domain, and Sec.~\ref{sec:physical-policy} introduces the physical-property-conditioned policy.

\vspace{-0.6em}
\subsection{Problem Formulation}
Our goal is to predict object mass logits $m\in\mathbb{R}^{3}$, friction logits $f\in\mathbb{R}^{3}$, and a continuous stiffness value $s\in\mathbb{R}$ from a synchronized visual window sequence $I_{\mathrm{v}}$ and a fingertip tactile image sequence $I_{\mathrm{t}}$, $B$ is the batch size. To capture object dynamics during grasping, we compute frame-to-frame flows from adjacent post-processed frames:
\begin{equation}
Flow_{\mathrm{v}}^{\tau}
=
\Phi_{\mathrm{flow}}
\left(I_{\mathrm{v}}^{\tau},I_{\mathrm{v}}^{\tau+1}\right),
Flow_{\mathrm{t}}^{\tau}
=
\Phi_{\mathrm{flow}}
\left(I_{\mathrm{t}}^{\tau},I_{\mathrm{t}}^{\tau+1}\right),
\end{equation}
where $\tau\in\{1,\ldots,14\}$ and $\Phi_{\mathrm{flow}}$ denotes the Farnebäck optical-flow operator~\cite{farneback}. The resulting visual and tactile flow sequences have dimensions
$F_{\mathrm{v}}$ and
$F_{\mathrm{t}}$, respectively.

\vspace{-0.6em}
\subsection{Visual-Tactile Encoders}
\label{sec:vt-encoders}
As shown in Fig.~\ref{figs:model1}(A), each modality has a content stream $E^c$ for sampled frames and a motion stream $E^m$ for flow fields. Both use ResNet-18~\cite{resnet} backbones followed by projection layers. The content feature at the first sampled frame is omitted when it is aligned with the 14 flow fields. Content and motion features are then concatenated and projected to form $Z_{\mathrm{v}},Z_{\mathrm{t}}\in\mathbb{R}^{B\times14\times512}$. Modality-specific GRUs return the hidden state at every time step, producing $F_{\mathrm{v}},F_{\mathrm{t}}\in\mathbb{R}^{B\times14\times512}$. Retaining the temporal axis allows the subsequent cross-attention layers to align visual and tactile interaction cues before temporal pooling.

\vspace{-0.6em}
\subsection{VLM-Derived Semantic Prior}
\label{sec:vlm-prior}
To complement contact-based observations, we generate a semantic prior from RGB frames acquired strictly before contact. GPT-5.4 analyzes the first five frames in the pre-contact buffer and describes only visible evidence, including object category, material, texture, geometry, and fill-state cues. It does not receive tactile frames, contact deformation, or post-contact motion. The prompt uses positive and negative in-context examples and requests confidence-aware hypotheses for mass, stiffness, and friction coefficient.

The generated text is encoded by frozen Sentence-BERT~\cite{sentencebert} to obtain a global feature $F_{\mathrm{sent}}\in\mathbb{R}^{B\times768}$ and by frozen BERT~\cite{bert} to obtain word-level features $F_{\mathrm{word}}\in\mathbb{R}^{B\times256\times768}$. We concatenate and precompute these encoder outputs offline as
$F_{\mathrm{text}}=\operatorname{Concat}[F_{\mathrm{sent}};F_{\mathrm{word}}]\in\mathbb{R}^{B\times257\times768}$. A learned projection and layer normalization then produce
\begin{equation}
F_{\mathrm{text}}
=
\operatorname{LN}
\left(
W_{\mathrm{text}}F_{\mathrm{text}}
\right)
\in\mathbb{R}^{B\times257\times256}.
\end{equation}
From a probabilistic perspective, the semantic prior conditions the posterior distribution of the physical properties:
\begin{equation}
P(\mathbf{y}\mid F_{\mathrm{v}},F_{\mathrm{t}},F_{\mathrm{text}})
\propto
P(F_{\mathrm{v}},F_{\mathrm{t}}\mid\mathbf{y},F_{\mathrm{text}})
P(\mathbf{y}\mid F_{\mathrm{text}}).
\label{eq:text_prior_bayes}
\end{equation}

Here, $\mathbf{y}=(m,s,f)$ denotes the joint physical-property target. This factorization motivates treating the pre-contact text as a prior and the post-contact visual--tactile sequences as interaction evidence; it is not an additional training objective.

\vspace{-0.6em}
\subsection{Multimodal Fusion and Prediction}
\label{sec:fusion}

ViTacPhys first performs bidirectional cross-attention between $F_{\mathrm{v}}$ and $F_{\mathrm{t}}$: visual queries attend to tactile keys and values, and tactile queries attend to visual keys and values. The two attended sequences are concatenated and temporally pooled to obtain $F_{\mathrm{vt}}\in\mathbb{R}^{B\times1024}$. This vector queries the VLM-derived token sequence in a multi-head cross-attention layer. A learned gate regulates the text contribution:
\begin{equation}
\begin{aligned}
CA_{\mathrm{text}} &= \operatorname{MHA}(F_{\mathrm{vt}}, F_{\mathrm{text}}, F_{\mathrm{text}}), \\
F_{\mathrm{fused}} &=
\operatorname{Concat}
\left[
F_{\mathrm{vt}};
\,
\sigma(W_g CA_{\mathrm{text}}) \odot CA_{\mathrm{text}}
\right],
\end{aligned}
\label{eq:gated_cross_attention}
\end{equation}
where $\operatorname{MHA}(Q,K,V)$ denotes multi-head cross-attention, $\sigma(\cdot)$ is the sigmoid function, and $W_g$ is a learnable gate projection. The direct visual--tactile branch preserves sensor evidence when the text gate is small. Finally, $F_{\mathrm{fused}}$ is passed to three task-specific heads. The mass and friction-coefficient heads each output one scalar ordinal logit, whereas the stiffness head predicts a normalized continuous scalar. In summary, ViTacPhys takes visual and tactile sequences as input, derives a VLM-based text modality from visual observations, and fuses the three modality-specific representations through cross-attention to predict object physical properties.

\vspace{-0.6em}
\subsection{Training Loss}
\label{sec:training-loss}
ViTacPhys jointly optimizes three physical properties prediction. The stiffness head is trained with mean-squared regression error. For mass and friction classification, the labels are discretized from continuous physical quantities and thus retain an inherent ordinal structure, such as low, medium, and high. Standard cross-entropy loss~\cite{crossentropy} treats classes as independent categories and does not model their relative ordering. We therefore adopt ordinal regression~\cite{ordinal} for mass and friction coefficient, denoted by $p\in{m,f}$. For each sample $i$, the corresponding prediction head outputs a scalar ordinal score $\eta_i^p$. Two learnable cutpoints, constrained by $\theta_0^p<\theta_1^p$, partition the score into three ordered classes. Let $\pi_i^p(k)$ denote the probability that sample $i$ belongs to class $k\in{0,1,2}$. The class probabilities are computed as
\begin{equation}
\pi_i^p(k)=
\begin{cases}
\sigma\left(\theta_0^p-\eta_i^p\right),
& k=0 \; (\eta_i^p < \theta_0^p),\\[2pt]
\sigma\left(\theta_1^p-\eta_i^p\right)
-\sigma\left(\theta_0^p-\eta_i^p\right),
& k=1 \; (\theta_0^p \leq \eta_i^p < \theta_1^p),\\[2pt]
1-\sigma\left(\theta_1^p-\eta_i^p\right),
& k=2 \; (\eta_i^p \geq \theta_1^p).
\end{cases}
\label{eq:ordinal_prob}
\end{equation}

Given the ground-truth ordinal label $y_i^p\in\{0,1,2\}$, the ordinal and stiffness losses are defined as
\begin{equation}
\mathcal{L}_{\mathrm{ord}}^p
=
-\frac{1}{B}\sum_{i=1}^{B}\log \pi_i^p(y_i^p),
\mathcal{L}_{\mathrm{stiff}}
=
\frac{1}{B}\sum_{i=1}^{B}(\hat{s}_i-s_i)^2.
\label{eq:task_losses}
\end{equation}
where $\hat{s}_i$ and $s_i$ denote the predicted and ground-truth stiffness values, respectively. The overall physical-property prediction loss is
\begin{equation}
\mathcal{L}_{\mathrm{phys}}
=
\lambda_m\mathcal{L}_{\mathrm{ord}}^m
+
\lambda_s\mathcal{L}_{\mathrm{stiff}}
+
\lambda_f\mathcal{L}_{\mathrm{ord}}^f.
\label{eq:phys_loss}
\end{equation}
We use GradNorm~\cite{gradnorm} to adaptively balance the task weights $\lambda_m$, $\lambda_s$, and $\lambda_f$. The gradient norms are measured at the final shared layer before the task-specific heads, promoting balanced optimization across the three prediction tasks.


\vspace{-0.6em}
\subsection{ViTacPhys for Downstream Adaptive Grasping Policy}
\label{sec:physical-policy-overview}
\subsubsection{Human-to-Robot Transfer}
\label{sec:human-to-robot}
ViTacPhys is trained on human demonstrations, but its deployment in downstream robotic tasks requires transferring the learned physical-property representations to our dexterous-hand platform. As described in Sec.~\ref{sec:dataset_system}, we first reduce the domain gap at the hardware level by using the same types of visual and tactile sensors in both the human data-acquisition system and the target robotic platform.

To reduce the visual domain gap, we condition Seedance 2.0 on robot-hand references and convert a subset of human demonstrations into robot-style grasping videos. As shown in Fig.~\ref{figs:model1}(B), the edited RGB videos remain paired with the original synchronized tactile observations. Inspired by human-to-robot scaling approaches~\cite{egoscale}, we also collect robot teleoperation data with motions matched to the human grasp prototypes. The visually edited demonstrations reuse the pre-contact VLM priors because their object identity and pre-contact scene content are unchanged. Robot teleoperation clips receive new descriptions from the same VLM pipeline. We fine-tune the human-pretrained predictor using robot teleoperation, matched-action human demonstrations, and visually edited human demonstrations. This procedure encourages cross-embodiment physical cues rather than embodiment-specific appearance.

\subsubsection{Policy Learning}
\label{sec:physical-policy}
We evaluate the ViTacPhys predictions in contact-rich adaptive grasping. A binary indicator $c\in\{0,1\}$ distinguishes pre-contact and post-contact control. Contact is detected when the tactile response within a temporal window changes beyond a preset threshold. Before contact, $c=0$ and the physical-property inputs are set to zero. Once contact is detected, the padded rolling queue described above enables an immediate first prediction, with the repeated entries progressively replaced by new observations.

Mass and friction coefficient are represented by predicted class indices $\hat{y}^m,\hat{y}^f\in\{1,2,3\}$. The continuous stiffness prediction is discretized into a bin index $\hat{y}^s\in\{0,\ldots,9\}$ using fixed quantile-based boundaries derived only from the ground-truth stiffness distribution. These boundaries are determined before downstream policy training and remain fixed during both downstream training and evaluation. The physical-property token is constructed as $\mathrm{t}_{\mathrm{phys}}=\operatorname{concat}\bigl(\operatorname{Emb}(c),\operatorname{Emb}(\hat{y}^m),\operatorname{Emb}(\hat{y}^s),\operatorname{Emb}(\hat{y}^f)\bigr)\in\mathbb{R}^{4\times 512}$. To reduce prediction fluctuations during real-world deployment, we apply temporal voting analogous to action chunking and pass the most probable prediction over recent windows for each property to the downstream policy. The resulting token is provided as a structured input to the ACT-style imitation-learning policy $M_{\mathrm{ACT}}$, as illustrated in Fig.~\ref{figs:model1}(C).

At policy time step $t$, the observation is $o_t=(t_{\mathrm{phys},t},x_t^{\mathrm{rgb}},p_t)$, where $x_t^{\mathrm{rgb}}$ is the wrist RGB frame and $p_t\in\mathbb{R}^{6}$ is the dexterous-hand proprioceptive state. The policy predicts dexterous-hand actions only; the arm is controlled separately and lies outside the learned action space. The policy is trained with the standard ACT~\cite{act} objective:
\vspace{-0.3em}
\begin{equation}
\mathcal{L}_{\mathrm{ACT}}
=
w_1 \mathcal{L}_{\mathrm{action}}
+
w_2 \mathcal{L}_{\mathrm{KL}},
\label{eq:act_loss}
\end{equation}
where $w_1$ and $w_2$ are loss weights, $\mathcal{L}_{\mathrm{action}}$ is the $\ell_1$ error between predicted and ground truth action sequences, and $\mathcal{L}_{\mathrm{KL}}$ regularizes the latent action-style distribution toward a Gaussian prior. The property token allows the policy to associate object-dependent mass, stiffness, and friction-coefficient classes with the demonstrated finger-closure and grasp-force profiles. More details are introduced in Sec.~\ref{sec:policy_learning_setup}.


\vspace{-0.6em}
\section{ViTacPhys Prediction Evaluation}
\label{sec:experiments}

\subsection{Experimental Setup}
\label{sec:experiments_setup}
\topic{Evaluation Protocol.}
We evaluate ViTacPhys under three data-split settings. Unless otherwise specified, $80\%/10\%/10\%$ is a target rather than an exact ratio. When an object-level split requires integer allocation within a category, we use the closest feasible allocation to an approximately $8{:}1{:}1$ ratio while retaining validation and test coverage. Each run uses an independent random seed for split assignment and model initialization, so the training, validation, and test files are resampled across runs.

(a) \emph{In-distribution}: all episodes from all objects are randomly split at the episode level. Thus, different grasping episodes of the same object may appear in the training, validation, and test sets. This setting evaluates ViTacPhys on new interactions with seen objects.

(b) \emph{Held-out-object}: the data are split at the object level within each coarse category, such that all episodes of an object belong to the same subset. Each coarse category contains at least one validation object and one test object. This setting evaluates generalization to unseen but semantically related objects from categories observed during training.

(c) \emph{One-shot}: under this extremely limited-data setting, we randomly select one object from each coarse category and use all of its episodes for training. All remaining objects are used for testing. This setting evaluates whether ViTacPhys can learn transferable physical-property representations from limited object diversity.

\topic{Metrics.}
For mass and friction-coefficient classification, we report accuracy and Macro-F1, defined as Macro-F1 $ = \frac{1}{C}\sum_{c=1}^{C}\frac{2P_cR_c}{P_c+R_c}$, where $C$ is the number of classes, and $P_c$ and $R_c$ denote the precision and recall of class $c$, respectively. Macro-F1 weights all classes equally and is therefore less sensitive to class imbalance.

For stiffness regression, we report mean absolute error (MAE), root mean squared error (RMSE), mean absolute percentage error (MAPE), the Pearson correlation coefficient $r$, and Within-$k$. The Pearson correlation coefficient between the predictions $\hat{y}$ and ground-truth values $y$ is defined as
\vspace{-0.3em}
\begin{equation}
r =
\frac{
\sum_i(\hat{y}_i-\overline{\hat{y}})(y_i-\bar{y})
}{
\sqrt{\sum_i(\hat{y}_i-\overline{\hat{y}})^2}
\sqrt{\sum_i(y_i-\bar{y})^2}
}.
\end{equation}

It measures the strength and direction of the linear relationship between predictions and ground truth, rather than their absolute agreement.
MAE measures the average absolute error, RMSE penalizes large errors more strongly, and MAPE measures the relative percentage error. Within-$k$ denotes the percentage of predictions whose relative error is below a tolerance $k$. Accuracy, MAE, and MAPE are reported as mean $\pm$ standard deviation over three training runs.

For each protocol and seed, every window from test set are included. Results are averaged over three independent runs and computed at the temporal-window level because both deployment and prediction operate on windows. Classification accuracy is episode-balanced: each window is weighted by the inverse of the number of evaluated windows in its episode, so longer episodes do not contribute more total weight. Window predictions are not collapsed to an object-level output because they can evolve over the interaction. The dataset-level mass and friction-coefficient class boundaries described in Sec.~\ref{sec:dataset} are defined before splitting and then fixed across all protocols and seeds.

\topic{Implementation Details.}
ViTacPhys is trained for $30$ epochs with a batch size of $32$. The VLM-derived text and the frozen Sentence-BERT/BERT encoder features $F_{\mathrm{text}}$ are generated offline. The learned text projection and all visual--tactile predictor parameters are trained end-to-end. We use AdamW~\cite{adamw} with an initial learning rate of $5\times10^{-5}$ and a weight decay of $10^{-4}$. For ordinal regression, the fixed sigmoid scale $\alpha$ in Eq.~\eqref{eq:ordinal_prob} is set to $15$. GradNorm is optimized with a learning rate of $5\times10^{-4}$ and asymmetry parameter $\alpha_{\mathrm{GN}}=1.5$.

\vspace{-0.6em}
\subsection{Physical-Property Prediction Results}
\label{sec:phys_results}

\begin{table*}[t]
\caption{Physical-property prediction results averaged over three independent training runs. The main evaluation metrics are highlighted with a gray background.}
\label{tab:phys_prediction}
\centering
\scriptsize
\renewcommand{\arraystretch}{0.98}
\setlength{\tabcolsep}{2.4pt}
\resizebox{\textwidth}{!}{%
\begin{tabular}{l>{\columncolor{gray!15}}c c >{\columncolor{gray!15}}c c >{\columncolor{gray!15}}c c c c >{\columncolor{gray!15}}c c}
\toprule
& \multicolumn{2}{c}{\textbf{Mass}} & \multicolumn{6}{c}{\textbf{Stiffness}} & \multicolumn{2}{c}{\textbf{Friction Coeff.}} \\
\cmidrule(lr){2-3}\cmidrule(lr){4-9}\cmidrule(lr){10-11}
\textbf{Setting} & \textbf{Acc. $\uparrow$} & \textbf{Macro-F1 $\uparrow$} & \textbf{MAE $\downarrow$} & \textbf{RMSE $\downarrow$} & \textbf{MAPE $\downarrow$} & \textbf{Pearson $r$ $\uparrow$} & \textbf{Within 10\% $\uparrow$} & \textbf{Within 20\% $\uparrow$} & \textbf{Acc. $\uparrow$} & \textbf{Macro-F1 $\uparrow$} \\
\midrule
In distribution & 0.972 $\pm$ 0.008 & 0.970 & 0.248 $\pm$ 0.032 & 0.386 & 5.51 $\pm$ 0.45 & 0.980 & 84.22 & 96.56 & 0.988 $\pm$ 0.005 & 0.984 \\
Held-out-object & 0.875 $\pm$ 0.019 & 0.866 & 0.435 $\pm$ 0.127 & 0.577 & 9.08 $\pm$ 2.31 & 0.947 & 61.56 & 89.64 & 0.975 $\pm$ 0.003 & 0.971 \\
One-shot & 0.492 $\pm$ 0.072 & 0.480 & 0.837 $\pm$ 0.045 & 1.457 & 16.84 $\pm$ 0.95 & 0.680 & 49.47 & 75.19 & 0.569 $\pm$ 0.089 & 0.527 \\
\bottomrule
\end{tabular}}
\vspace{-0.8em}
\end{table*}

Table~\ref{tab:phys_prediction} summarizes ViTacPhys under the three evaluation protocols. In distribution, mass and friction-coefficient accuracies exceed $97\%$, stiffness MAPE is $5.51\%$, and $96.56\%$ of stiffness predictions fall within $20\%$ relative error. The mass and friction-coefficient Macro-F1 scores are $0.970$ and $0.984$, respectively, and stiffness reaches a Pearson correlation of $r=0.980$. These results show consistent predictions for new grasp interactions with objects represented during training.

The held-out-object setting evaluates generalization to unseen objects within known coarse categories. ViTacPhys achieves $87.5\%$ mass accuracy, $97.5\%$ friction-coefficient accuracy, a stiffness MAPE of $9.08\%$, and a stiffness Pearson correlation of $r=0.947$. Under the more challenging one-shot setting, mass and friction-coefficient accuracies decrease to $49.2\%$ and $56.9\%$, respectively, while stiffness MAPE increases to $16.84\%$. These results show that generalization degrades substantially when training-object diversity is severely limited.


\subsection{Ablation Studies}
\label{sec:ablations}

\begin{table}[!t]
\vspace{0.6em}
\caption{Input-modality ablation. Bold and underlined entries indicate the best and second-best results. ``w/o'' denotes without, while ``w'' denotes with.}
\label{tab:modality_ablation}
\centering
\tiny
\renewcommand{\arraystretch}{0.82}
\setlength{\tabcolsep}{1.8pt}
\resizebox{\columnwidth}{!}{%
\begin{tabular}{lcccccccccc}
\toprule
& \multicolumn{2}{c}{\textbf{Mass}} & \multicolumn{6}{c}{\textbf{Stiffness}} & \multicolumn{2}{c}{\textbf{Friction Coeff.}} \\
\cmidrule(lr){2-3}\cmidrule(lr){4-9}\cmidrule(lr){10-11}
\textbf{Experiment} & \textbf{Acc. $\uparrow$} & \textbf{F1 $\uparrow$} & \textbf{MAE $\downarrow$} & \textbf{RMSE $\downarrow$} & \textbf{MAPE $\downarrow$} & \textbf{Pearson $r$ $\uparrow$} & \textbf{W10 $\uparrow$} & \textbf{W20 $\uparrow$} & \textbf{Acc. $\uparrow$} & \textbf{F1 $\uparrow$} \\
\midrule
w/o tactile content, w text & 0.660 & 0.648 & 0.756 & 1.152 & 16.70 & 0.769 & 46.39 & 80.69 & 0.915 & 0.905 \\
w/o visual content, w text & 0.660 & 0.640 & 0.668 & 0.853 & 13.12 & 0.887 & 45.97 & 79.79 & 0.844 & 0.862 \\
w/o visual flow, w text & 0.757 & 0.751 & 0.457 & 0.579 & 9.44 & 0.942 & 59.70 & 88.39 & 0.905 & 0.886 \\
w/o tactile flow, w text & 0.736 & 0.727 & \textbf{0.412} & \textbf{0.563} & \textbf{8.54} & \textbf{0.950} & \textbf{70.87} & \underline{89.52} & 0.888 & 0.870 \\
w/o both flow, w text & 0.661 & 0.662 & 0.858 & 1.366 & 18.91 & 0.633 & 56.94 & 73.47 & 0.898 & 0.857 \\
Text only & 0.542 & 0.525 & 0.769 & 1.943 & 25.37 & 0.580 & 42.37 & 70.00 & 0.742 & 0.740 \\
w/o text & \underline{0.810} & \underline{0.827} & 0.630 & 0.796 & 12.43 & 0.900 & 56.46 & 77.36 & \textbf{0.981} & \textbf{0.978} \\
Ours & \textbf{0.875} & \textbf{0.866} & \underline{0.435} & \underline{0.577} & \underline{9.08} & \underline{0.947} & \underline{61.56} & \textbf{89.64} & \underline{0.975} & \underline{0.971} \\
\bottomrule
\end{tabular}}
\vspace{0.2em}
\end{table}

The held-out-object protocol evaluates generalization to unseen objects based on knowledge learned from seen objects. We therefore use this setting for the following ablation studies.

\topic{Input Modalities.}
Table~\ref{tab:modality_ablation} studies the visual, tactile, and textual inputs and the two flow branches. Removing either the visual or tactile content stream reduces mass accuracy by up to $21.5$ percentage points and increases stiffness MAPE by up to $7.62$ percentage points. Removing the visual content stream causes the largest decrease in friction-coefficient accuracy, from $97.5\%$ to $84.4\%$. These results indicate that the two content streams provide complementary cues.

The flow ablation reveals a task-dependent trade-off. Removing tactile flow reduces mass and friction-coefficient accuracy by $13.9$ and $8.7$ percentage points, respectively, but improves stiffness MAPE from $9.08\%$ to $8.54\%$ and Within-$10\%$ from $61.56\%$ to $70.87\%$. Removing both flow branches degrades all three tasks. Thus, flow is important overall, but the tactile-flow branch does not consistently improve stiffness regression.

Using the textual prior alone performs poorly, including a mass accuracy of $54.2\%$, confirming that text cannot replace interaction observations. Removing text decreases mass accuracy from $87.5\%$ to $81.0\%$ and increases stiffness MAPE from $9.08\%$ to $12.43\%$, but slightly increases friction-coefficient accuracy from $97.5\%$ to $98.1\%$. The VLM prior therefore complements visual--tactile evidence for mass and stiffness, but does not improve every metric.

\begin{table}[!t]
\vspace{0.6em}
\caption{Ablation of multi-task loss weighting strategies and loss functions.}
\label{tab:loss_ablation}
\centering
\tiny
\renewcommand{\arraystretch}{0.82}
\setlength{\tabcolsep}{1.8pt}
\resizebox{\columnwidth}{!}{%
\begin{tabular}{lcccccccccc}
\toprule
& \multicolumn{2}{c}{\textbf{Mass}} & \multicolumn{6}{c}{\textbf{Stiffness}} & \multicolumn{2}{c}{\textbf{Friction Coeff.}} \\
\cmidrule(lr){2-3}\cmidrule(lr){4-9}\cmidrule(lr){10-11}
\textbf{Experiment} & \textbf{Acc. $\uparrow$} & \textbf{F1 $\uparrow$} & \textbf{MAE $\downarrow$} & \textbf{RMSE $\downarrow$} & \textbf{MAPE $\downarrow$} & \textbf{Pearson $r$ $\uparrow$} & \textbf{W10 $\uparrow$} & \textbf{W20 $\uparrow$} & \textbf{Acc. $\uparrow$} & \textbf{F1 $\uparrow$} \\
\midrule
Manual loss weights & 0.778 & 0.799 & 0.540 & 0.714 & 11.01 & 0.922 & 59.51 & 83.06 & 0.935 & 0.938 \\
DWA loss weights & 0.762 & 0.772 & 0.576 & 0.764 & 12.38 & 0.880 & 54.72 & 84.44 & 0.947 & 0.950 \\
Cross entropy & 0.635 & 0.657 & 0.534 & 0.730 & 10.07 & 0.945 & 59.38 & 85.14 & \textbf{0.979} & \textbf{0.978} \\
Ours & \textbf{0.875} & \textbf{0.866} & \textbf{0.435} & \textbf{0.577} & \textbf{9.08} & \textbf{0.947} & \textbf{61.56} & \textbf{89.64} & \underline{0.975} & \underline{0.971} \\
\bottomrule
\end{tabular}}
\vspace{0.2em}
\end{table}

\topic{Loss Weighting.}
Table~\ref{tab:loss_ablation} compares multi-task loss-weighting strategies. Manual weighting assigns equal weights to the three tasks, whereas Dynamic Weight Averaging (DWA)~\cite{dwa} adjusts them from the relative changes in task losses. DWA provides no consistent improvement over manual weighting in our setting. GradNorm~\cite{gradnorm} balances gradient magnitudes at the shared encoder and, relative to manual weighting, improves mass and friction-coefficient accuracy by $9.7$ and $4.0$ percentage points and reduces stiffness MAPE by $1.93$ percentage points.

\topic{Ordinal Loss.}
Table~\ref{tab:loss_ablation} compares ordinal regression with standard cross-entropy loss~\cite{crossentropy}. Cross entropy treats the classes independently and ignores their ordering. Replacing ordinal regression with cross entropy reduces mass accuracy from $87.5\%$ to $63.5\%$, indicating that the ordinal structure is important for mass. For friction coefficient, however, cross entropy is $0.4$ percentage points more accurate than ordinal regression ($97.9\%$ versus $97.5\%$), so the ordinal objective does not improve both classification tasks.

\vspace{-0.6em}
\section{Downstream Evaluation in Adaptive Grasping}

We investigate whether an imitation-learning policy can benefit from the physical-property predictions provided by ViTacPhys. The adaptive grasping policy is introduced in Sec.~\ref{sec:physical-policy}.


\subsection{Real-World Experimental Setup}
\label{sec:policy_learning_setup}

\begin{figure}[!t]
\centering
\includegraphics[width=.92\linewidth]{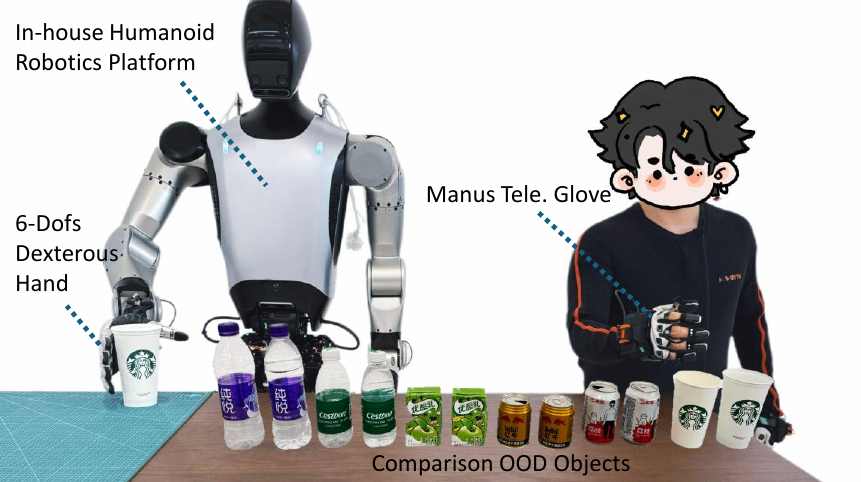}
\vspace{-0.6em}
   \caption{Robot teleoperation setup for downstream grasping. A Manus glove controls the dexterous hand to collect demonstrations in which the robot grasps and lifts objects with property-dependent forces.}
\label{figs:teleoperation}
\vspace{-0.3em}
\end{figure}

\topic{Platform.}
The downstream evaluations are conducted on an in-house dexterous robotic platform consisting of a 7-DoF manipulator and a linkage-driven 6-DoF dexterous hand. As described in Sec.~\ref{sec:dataset_system}, the hand uses the same visual and tactile sensor configuration, including sensor type and tactile spatial density, as the human data-collection system. We collect robot teleoperation demonstrations using a Quantum Manus glove, as shown in Fig.~\ref{figs:teleoperation}, following the same vertical-grasping protocol used for the human demonstrations.

\topic{Data Splits.}
We select $40$ robot-graspable objects from the ViTacPhys dataset as the in-distribution (ID) set. To evaluate transfer beyond the held-out-object setting in Sec.~\ref{sec:experiments_setup}, the out-of-distribution (OOD) set contains six pairs of visually similar objects with different measured properties, as shown in Fig.~\ref{figs:teleoperation}. None of these OOD objects is used during human pretraining or robot-policy training. We collect $10$ expert teleoperation demonstrations per object. The operator applies stronger grip forces to objects with lower friction coefficients while avoiding excessive deformation of soft objects.

ViTacPhys is fine-tuned on $80\%$ of the ID teleoperation data together with matched-action human demonstrations and the visually augmented data described in Sec.~\ref{sec:human-to-robot}. It is evaluated on the remaining ID episodes and all OOD objects. The physical-property-conditioned policy $M_{\mathrm{phys}}$ is trained on all ID teleoperation demonstrations and evaluated in real-world grasping trials.

\topic{Metrics.}
We evaluate human-to-robot transfer using the same metrics as ViTacPhys. For downstream adaptive grasping, we report the grasp success rate and the similarity between robot and human teleoperated grasping forces. A trial is considered a clean success if the hand grasps and holds the object for more than $5$ seconds without dropping it. For deformable objects, clean success additionally requires no visible crushing or excessive deformation, whereas for rigid objects, successful lifting and holding is sufficient. If a deformable object is successfully lifted and held but visibly compressed due to excessive grasping force, the trial is recorded as an over-force success. This outcome is judged by the same evaluator through visual inspection, and only clearly visible deformation is counted as over-force. Each method is evaluated over three trials per object.

For force analysis, we retain clean-success trials and compute each trial's steady-state three-finger grasp signal $\bar{g}$ over a final $3$-s interval in which the tactile readings vary minimally. The teleoperation reference $\bar{g}^{h}_i$ is the mean over the expert demonstrations for object $i$. An object enters a two-method comparison when both methods have at least one clean-success trial on it. As defined in Sec.~\ref{sec:dataset}, the grasp signal is the sum of the normalized thumb, index, and middle-finger readings; Fig.~\ref{figs:force_trend} instead shows the corresponding sum of calibrated forces in newtons.

For the retained robot trials, the mean absolute force error is $E_{\mathrm{force}}=\frac{1}{N}\sum_{j=1}^{N}|\bar{g}^{r}_j-\bar{g}^{h}_{o(j)}|$, where $N$ is the number of retained trials and $o(j)$ denotes the object in trial $j$. For Pearson $r$ and pairwise force ordering, multiple clean-success trials are first averaged within each object. Pairwise comparisons whose teleoperation or robot values are tied are excluded. These metrics measure object-dependent force adaptation rather than only average force magnitude.

\topic{Real-time Deployment.}
ViTacPhys and $M_{\mathrm{phys}}$ are trained independently and combined only during deployment. The policy is trained using ground-truth physical-property indices with zero-mean Gaussian perturbations, whereas deployment supplies the cumulative-voted predictions from ViTacPhys. The control loop runs at $30$~Hz, and both networks are exported to ONNX. On an NVIDIA Jetson Orin, ViTacPhys and the policy require $9$~ms and $10$~ms per inference, respectively.

Before contact, the VLM processes five buffered RGB frames to generate the semantic prior. This one-time initialization takes approximately $10$~s and is completed before the grasp begins. At contact onset, unavailable entries in the synchronized 30-frame visual--tactile queue repeat the earliest post-contact observation. ViTacPhys therefore produces an initial property estimate immediately; the padded entries are progressively replaced and rolling predictions continue as new frames arrive. The $30$~Hz rate refers to sensing, rolling prediction, and policy control, not to the one-time VLM call.

\topic{Baselines.}
For downstream adaptive grasping, we compare with ACT~\cite{act} and ViTacFormer~\cite{2025vitacformer}. All policies use a common transformer backbone with four encoder and four decoder layers and the same optimization budget; their method-specific inputs and objectives remain unchanged. During training of $M_{\mathrm{phys}}$, zero-mean Gaussian noise is added to its ground-truth physical-property index inputs before embedding to improve tolerance to prediction fluctuations.

\vspace{-0.6em}
\subsection{Human-to-Robot Transfer Results}
\label{sec:transfer-results}
\begin{table}[t]
\caption{Human-to-robot transfer. Bold and underlined entries denote the best and second-best results.}
\label{tab:transfer_prediction}
\centering
\tiny
\renewcommand{\arraystretch}{0.66}
\setlength{\tabcolsep}{0.55pt}
\resizebox{\columnwidth}{!}{%
\begin{tabular}{@{}ccccccccccc@{}}
\toprule
\multirow[c]{2}{*}{\textbf{Split}} & \multirow[c]{2}{*}{\textbf{Init.}} & \multicolumn{3}{c}{\textbf{Training Data}} & \multicolumn{2}{c}{\textbf{Mass}} & \multicolumn{2}{c}{\textbf{Stiffness}} & \multicolumn{2}{c}{\textbf{Friction Coeff.}} \\
\cmidrule(lr){3-5}\cmidrule(lr){6-7}\cmidrule(lr){8-9}\cmidrule(lr){10-11}
& & \textbf{Tele.} & \textbf{Aug.} & \textbf{H} & \textbf{Acc. $\uparrow$} & \textbf{Macro-F1 $\uparrow$} & \textbf{MAE $\downarrow$} & \textbf{MAPE $\downarrow$} & \textbf{Acc. $\uparrow$} & \textbf{Macro-F1 $\uparrow$} \\
\midrule
\multirow[c]{6}{*}[-3.5pt]{\textbf{ID}} & \multirow[c]{2}{*}{Train from scratch}
& $\checkmark$ & & & \textbf{0.998} & \textbf{0.996} & 0.217 & 5.10 & \textbf{1.000} & \textbf{1.000} \\
& & $\checkmark$ & $\checkmark$ & $\checkmark$ & \underline{0.960} & \underline{0.966} & 0.315 & 6.78 & 0.965 & 0.973 \\
\cmidrule(lr){2-11}
& \multirow[c]{4}{*}{Fine-tuned}
& $\checkmark$ & & & 0.907 & 0.912 & 0.201 & 4.66 & 0.949 & 0.960 \\
& & $\checkmark$ & & $\checkmark$ & 0.926 & 0.927 & \textbf{0.164} & \textbf{3.66} & \underline{0.986} & \underline{0.989} \\
& & $\checkmark$ & $\checkmark$ & & 0.922 & 0.908 & \underline{0.171} & \underline{3.97} & 0.950 & 0.962 \\
& & $\checkmark$ & $\checkmark$ & $\checkmark$ & 0.910 & 0.917 & 0.203 & 4.34 & 0.962 & 0.971 \\
\midrule
\multirow[c]{6}{*}[-3.5pt]{\textbf{OOD}} & \multirow[c]{2}{*}{Train from scratch}
 & $\checkmark$ & & & 0.445 & 0.340 & 1.268 & 35.90 & 0.587 & 0.443 \\
& & $\checkmark$ & $\checkmark$ & $\checkmark$ & 0.412 & 0.454 & 1.596 & 42.45 & 0.687 & 0.639 \\
\cmidrule(lr){2-11}
& \multirow[c]{4}{*}{Fine-tuned}
 & $\checkmark$ & & & 0.544 & 0.511 & \underline{0.781} & \underline{20.29} & \underline{0.756} & \underline{0.722} \\
& & $\checkmark$ & & $\checkmark$ & 0.609 & 0.656 & 0.850 & 21.07 & 0.654 & 0.647 \\
& & $\checkmark$ & $\checkmark$ & & \underline{0.639} & \underline{0.660} & 0.837 & 21.54 & 0.693 & 0.650 \\
& & $\checkmark$ & $\checkmark$ & $\checkmark$ & \textbf{0.780} & \textbf{0.752} & \textbf{0.766} & \textbf{19.01} & \textbf{0.827} & \textbf{0.822} \\
\bottomrule
\end{tabular}%
}
\vspace{-0.8em}
\end{table}

For human-to-robot transfer, we initialize ViTacPhys from the in-distribution model described in Sec.~\ref{sec:experiments_setup}, which is pretrained on $80\%$ of the ViTacPhys episodes. Table~\ref{tab:transfer_prediction} compares different fine-tuning strategies using robot teleoperation data (Tele.), matched-action human demonstrations (H), and augmented dexterous-hand demonstrations (Aug.). On ID episodes, training ViTacPhys from scratch with $80\%$ of the teleoperation data achieves the best same-domain performance, while models initialized from the pretrained ViTacPhys and fine-tuned with different data combinations also achieve strong results, with mass and friction-coefficient accuracies above $90\%$ and stiffness MAPE below $5\%$.

The differences are more pronounced on OOD objects with similar appearance but different physical properties. With all three data sources, pretraining followed by fine-tuning improves mass and friction-coefficient accuracy over training from scratch by $36.8$ and $14.0$ percentage points, respectively, and reduces stiffness MAPE by $23.44$ percentage points.

Among the fine-tuned models, adding augmented data to teleoperation data improves mass accuracy from $54.4\%$ to $63.9\%$, but decreases friction-coefficient accuracy from $75.6\%$ to $69.3\%$ and increases stiffness MAPE from $20.29\%$ to $21.54\%$. Adding matched-action human data to this combination produces the best overall OOD result: $78.0\%$ mass accuracy, $19.01\%$ stiffness MAPE, and $82.7\%$ friction-coefficient accuracy. Thus, augmentation alone is not uniformly beneficial; the strongest transfer is obtained when human pretraining, robot teleoperation, matched-action human data, and robot-style visual augmentation are combined.

\vspace{-0.6em}
\subsection{Adaptive Grasping Results}
\label{sec:adaptive-grasp-results}

\begin{figure}[!t]
\centering
\includegraphics[width=.99\linewidth]{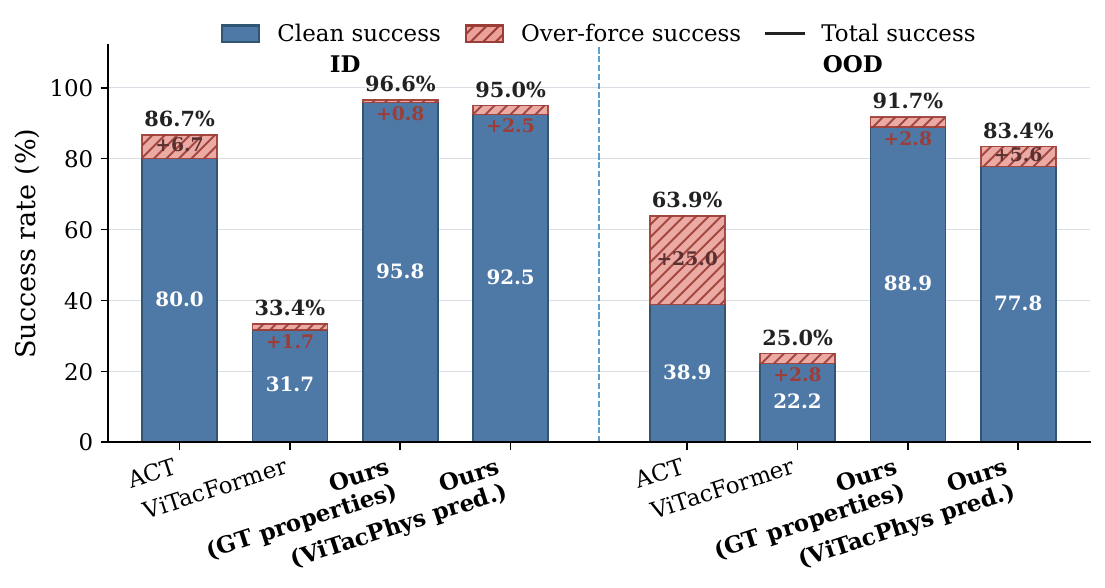}
\vspace{-1.6em}
\caption{Adaptive grasping outcomes on ID and OOD objects. ``GT Properties'' conditions our policy on ground-truth properties, whereas ``ViTacPhys Pred.'' uses predicted properties. Clean success is the primary metric. Total success is the sum of clean and over-force outcomes and is reported as lift-and-hold success.}
\label{figs:real_success}
\end{figure}

\begin{table}[!t]
\caption{Normalized grasping force comparison.}
\label{tab:grasp_force}
\centering
\tiny
\renewcommand{\arraystretch}{0.98}
\setlength{\tabcolsep}{2pt}
\resizebox{\columnwidth}{!}{%
\begin{tabular}{llccccc}
\toprule
\textbf{Comparison} & \textbf{Split} & \textbf{Method} & \textbf{Evaluated-object ratio} & \textbf{Abs. Err. $\downarrow$} & \textbf{Pearson $r$ $\uparrow$} & \textbf{Pairwise $\uparrow$} \\
\midrule
\multirow{4}{*}{ACT vs. Ours}
& \multirow{2}{*}{ID} & ACT~\cite{act} & \multirow{2}{*}{$37/40$ (92.5\%)} & 0.040 & \textbf{0.889} & \textbf{0.852} \\
& & Ours & & \textbf{0.035} & 0.862 & 0.843 \\
\cmidrule(lr){2-7}
& \multirow{2}{*}{OOD} & ACT~\cite{act} & \multirow{2}{*}{$8/12$ (66.7\%)} & 0.076 & 0.438 & 0.639 \\
& & Ours & & \textbf{0.061} & \textbf{0.852} & \textbf{0.806} \\
\midrule
\multirow{4}{*}{ViTacFormer vs. Ours}
& \multirow{2}{*}{ID} & ViTacFormer~\cite{2025vitacformer} & \multirow{2}{*}{$17/40$ (42.5\%)} & 0.077 & 0.503 & 0.733 \\
& & Ours & & \textbf{0.044} & \textbf{0.851} & \textbf{0.844} \\
\cmidrule(lr){2-7}
& \multirow{2}{*}{OOD} & ViTacFormer~\cite{2025vitacformer} & \multirow{2}{*}{$4/12$ (33.3\%)} & 0.116 & 0.637 & 0.667 \\
& & Ours & & \textbf{0.065} & \textbf{0.915} & \textbf{0.833} \\
\bottomrule
\end{tabular}}
\vspace{-0.8em}
\end{table}

Fig.~\ref{figs:real_success} shows the real-robot adaptive grasping results. ViTacFormer achieves the lowest success rates and fails on most ID and OOD objects. Compared with ACT, the policy conditioned on ViTacPhys predictions is associated with clean-success rates $12.5$ percentage points higher on ID objects and $38.9$ percentage points higher on OOD objects. Its total-success rates are $8.3$ and $19.5$ percentage points higher, respectively. Replacing the predictions with ground-truth properties is associated with a further total-success increase of $1.6$ percentage points on ID objects and $8.3$ percentage points on OOD objects.

The drop in clean success from ID to OOD is $14.7$ percentage points for our method with ViTacPhys predictions, compared with $41.1$ percentage points for ACT. Moreover, ACT produces over-force successes more frequently, indicating that a policy without explicit physical-property conditioning may rely more heavily on excessive grasping force. Representative failure cases are shown in Fig.~\ref{fig:teaser}.

\vspace{-0.6em}
\subsection{Similarity with Human Grasping Behavior}
To compare adaptive grasping behavior, we measure the similarity between deployed robot forces and human teleoperation. Table~\ref{tab:grasp_force} computes force metrics only on objects successfully grasped by both methods in each comparison and reports the proportion of objects retained. On ID objects, ACT and our method produce similar force profiles. On the common successful OOD objects, our method reduces absolute force error by $19.7\%$ relative to ACT and improves Pearson $r$ and pairwise ranking by $0.414$ and $0.167$, respectively. Because these metrics exclude failures, they describe force adaptation on the retained subset rather than all evaluated objects.

\begin{figure}[!t]
\centering
\includegraphics[width=.98\linewidth]{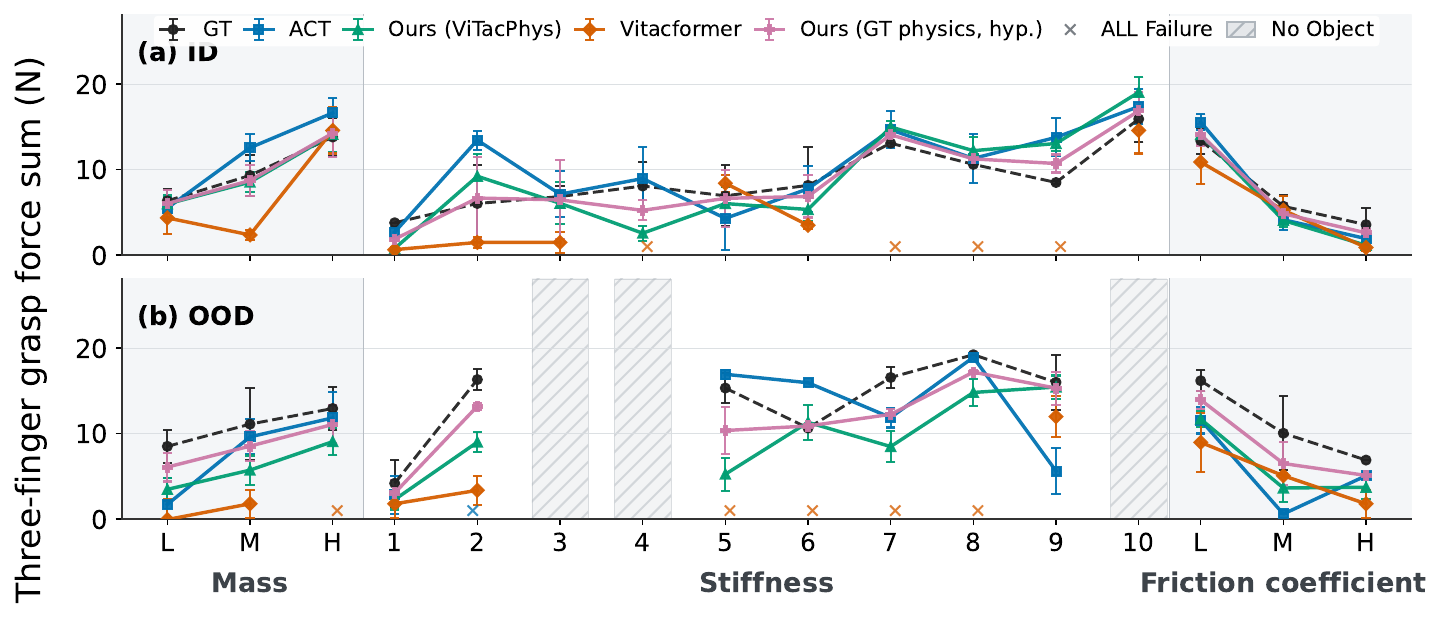}
\vspace{-0.6em}
\caption{Grasping force comparison between our method and baselines, grouped by the ground-truth mass, stiffness, and friction coefficient bins used as physical-property inputs to the downstream policy. ``L'', ``M'', and ``H'' denote low, medium, and high, respectively. For a given method, ``X'' indicates that all objects in the corresponding bin failed to be grasped by that method, while an empty bin indicates that no evaluated object in the corresponding split/subset belongs to that bin.}

\label{figs:force_trend}
\vspace{-0.3em}
\end{figure}

Fig.~\ref{figs:force_trend} compares grasping force across ground-truth property bins. Human teleoperation generally uses larger forces for heavier and stiffer objects and smaller forces for objects with higher friction coefficients. On the evaluated successful objects, our method follows these trends more closely than the baselines, although individual bins remain noisy.

\vspace{-1.6em}

\vspace{0.8em}
\section{Conclusion and Limitations}
\label{sec:limitation}
We presented ViTacPhys, a temporal visual--tactile framework that learns manipulation-relevant physical properties from human grasping demonstrations. ViTacPhys combines interaction cues with a VLM-derived semantic prior to predict mass class, stiffness, and friction-coefficient class. We transferred the human-trained predictor to a dexterous robot using limited teleoperation data and conditioned an adaptive grasping policy on its outputs. Relative to ACT, the conditioned policy was associated with clean-success rates $12.5$ percentage points higher on in-distribution objects and $38.9$ percentage points higher on visually similar but physically different out-of-distribution objects. On the common successful OOD objects, the resulting force profile was closer to human teleoperation. These results position ViTacPhys as a dataset, complete system, and feasibility study for connecting multimodal human demonstrations with adaptive robot control.

The system remains limited by sensing, data scale, and latency. The tactile maps mainly capture normal pressure, so mass and the silicone-contact friction coefficient are discretized. The stiffness target is an operational grasp-level measurement that includes object deformation and contact-system compliance rather than an intrinsic material constant. The dataset contains $60$ objects manipulated by one participant, and the OOD robot evaluation contains six object pairs with three trials per object.  Finally, deployment requires a one-time pre-contact VLM call, although padded post-contact queues allow physical-property prediction to begin immediately at contact. Future work will add richer force sensing, more participants and objects, broader OOD trials, and faster semantic priors.

\bibliographystyle{IEEEtran}
\bibliography{reference}

@book{ordinal,
  title={Ordinal regression models},
  author={Williams, Richard A and Quiroz, Christopher},
  year={2020},
  publisher={SAGE Publications Limited Thousand Oaks, CA}
}

@inproceedings{gradnorm,
  title={Gradnorm: Gradient normalization for adaptive loss balancing in deep multitask networks},
  author={Chen, Zhao and Badrinarayanan, Vijay and Lee, Chen-Yu and Rabinovich, Andrew},
  booktitle={International conference on machine learning},
  pages={794--803},
  year={2018},
  organization={PMLR}
}

@inproceedings{dwa,
  title={End-to-end multi-task learning with attention},
  author={Liu, Shikun and Johns, Edward and Davison, Andrew J},
  booktitle={Proceedings of the IEEE/CVF conference on computer vision and pattern recognition},
  pages={1871--1880},
  year={2019}
}

@article{kmeans,
  title={Algorithm AS 136: A k-means clustering algorithm},
  author={Hartigan, John A and Wong, Manchek A},
  journal={Journal of the royal statistical society. series c (applied statistics)},
  volume={28},
  number={1},
  pages={100--108},
  year={1979},
  publisher={JSTOR}
}

@inproceedings{farneback,
  title={Two-frame motion estimation based on polynomial expansion},
  author={Farneb{\"a}ck, Gunnar},
  booktitle={Scandinavian conference on Image analysis},
  pages={363--370},
  year={2003},
  organization={Springer}
}

@inproceedings{sentencebert,
  title={Sentence-bert: Sentence embeddings using siamese bert-networks},
  author={Reimers, Nils and Gurevych, Iryna},
  booktitle={Proceedings of the 2019 conference on empirical methods in natural language processing and the 9th international joint conference on natural language processing (EMNLP-IJCNLP)},
  pages={3982--3992},
  year={2019}
}

@inproceedings{bert,
  title={Bert: Pre-training of deep bidirectional transformers for language understanding},
  author={Devlin, Jacob and Chang, Ming-Wei and Lee, Kenton and Toutanova, Kristina},
  booktitle={Proceedings of the 2019 conference of the North American chapter of the association for computational linguistics: human language technologies, volume 1 (long and short papers)},
  pages={4171--4186},
  year={2019}
}

@inproceedings{crossentropy,
  title={Cross-entropy loss functions: Theoretical analysis and applications},
  author={Mao, Anqi and Mohri, Mehryar and Zhong, Yutao},
  booktitle={International conference on Machine learning},
  pages={23803--23828},
  year={2023},
  organization={pmlr}
}

@article{adamw,
  title={Decoupled weight decay regularization},
  author={Loshchilov, Ilya and Hutter, Frank},
  journal={International Conference on Learning Representations},
  year={2019}
}

@article{liu2025does,
  title={Does feasibility matter? understanding the impact of feasibility on synthetic training data},
  author={Liu, Yiwen and Bader, Jessica and Kim, Jae Myung},
  journal={arXiv preprint arXiv:2505.10551},
  year={2025}
}

@inproceedings{wimboeck2006passivity,
  title={Passivity-based object-level impedance control for a multifingered hand},
  author={Wimboeck, Thomas and Ott, Christian and Hirzinger, Gerhard},
  booktitle={2006 IEEE/RSJ International Conference on Intelligent Robots and Systems},
  pages={4621--4627},
  year={2006},
  organization={IEEE}
}

@inproceedings{impedence,
  title={Learning object-level impedance control for robust grasping and dexterous manipulation},
  author={Li, Miao and Yin, Hang and Tahara, Kenji and Billard, Aude},
  booktitle={2014 IEEE International Conference on Robotics and Automation (ICRA)},
  pages={6784--6791},
  year={2014},
  organization={IEEE}
}

@article{pi_05,
  title={pi $\{$0.5$\}$ : a Vision-Language-Action Model with Open-World Generalization},
  author={Intelligence, Physical and Black, Kevin and Brown, Noah and Darpinian, James and Dhabalia, Karan and Driess, Danny and Esmail, Adnan and Equi, Michael and Finn, Chelsea and Fusai, Niccolo and others},
  journal={arXiv preprint arXiv:2504.16054},
  year={2025}
}

@article{dreamzero,
  title={World action models are zero-shot policies},
  author={Ye, Seonghyeon and Ge, Yunhao and Zheng, Kaiyuan and Gao, Shenyuan and Yu, Sihyun and Kurian, George and Indupuru, Suneel and Tan, You Liang and Zhu, Chuning and Xiang, Jiannan and others},
  journal={arXiv preprint arXiv:2602.15922},
  year={2026}
}

@article{twostream,
  title={Two-stream network-driven vision-based tactile sensor for object feature extraction and fusion perception},
  author={Huang, Muxing and Chen, Zibin and Xu, Weiliang and Li, Zilan and Zhou, Yuanzhi and Zhou, Guoyuan and Chen, Wenjing and Li, Xinming},
  journal={arXiv preprint arXiv:2510.12528},
  year={2025}
}

@inproceedings{phys101,
  title={Physics 101: Learning Physical Object Properties from Unlabeled Videos.},
  author={Wu, Jiajun and Lim, Joseph J and Zhang, Hongyi and Tenenbaum, Joshua B and Freeman, William T},
  booktitle={BMVC},
  volume={2},
  number={6},
  pages={7},
  year={2016}
}

@article{sundaram2019learning,
  title={Learning the signatures of the human grasp using a scalable tactile glove},
  author={Sundaram, Subramanian and Kellnhofer, Petr and Li, Yunzhu and Zhu, Jun-Yan and Torralba, Antonio and Matusik, Wojciech},
  journal={Nature},
  volume={569},
  number={7758},
  pages={698--702},
  year={2019},
  publisher={Nature Publishing Group UK London}
}

@inproceedings{omnipush,
  title={Omnipush: accurate, diverse, real-world dataset of pushing dynamics with rgb-d video},
  author={Bauza, Maria and Alet, Ferran and Lin, Yen-Chen and Lozano-P{\'e}rez, Tom{\'a}s and Kaelbling, Leslie P and Isola, Phillip and Rodriguez, Alberto},
  booktitle={2019 IEEE/RSJ International Conference on Intelligent Robots and Systems (IROS)},
  pages={4265--4272},
  year={2019},
  organization={IEEE}
}

@article{force,
  title={Force: Dataset and method for intuitive physics guided human-object interaction},
  author={Zhang, Xiaohan and Bhatnagar, Bharat Lal and Starke, Sebastian and Petrov, Ilya and Guzov, Vladimir and Dhamo, Helisa and P{\'e}rez-Pellitero, Eduardo and Pons-Moll, Gerard},
  journal={CoRR},
  year={2024}
}

@inproceedings{yuan2017connecting,
  title={Connecting look and feel: Associating the visual and tactile properties of physical materials},
  author={Yuan, Wenzhen and Wang, Shaoxiong and Dong, Siyuan and Adelson, Edward},
  booktitle={Proceedings of the IEEE conference on computer vision and pattern recognition},
  pages={5580--5588},
  year={2017}
}

@inproceedings{synesthesia,
  title={Teaching cameras to feel: Estimating tactile physical properties of surfaces from images},
  author={Purri, Matthew and Dana, Kristin},
  booktitle={European Conference on Computer Vision},
  pages={1--20},
  year={2020},
  organization={Springer}
}

@inproceedings{depth2mass,
  title={Estimating Object Physical Properties from RGB-D Vision and Depth Robot Sensors Using Deep Learning},
  author={Cardoso, Ricardo Pedreiras and Moreno, Plinio},
  booktitle={Iberian Conference on Pattern Recognition and Image Analysis},
  pages={97--110},
  year={2025},
  organization={Springer}
}

@article{motion_kinematics,
  title={Predicting object properties based on movement kinematics},
  author={Kopnarski, Lena and Lippert, Laura and Rudisch, Julian and Voelcker-Rehage, Claudia},
  journal={Brain Informatics},
  volume={10},
  number={1},
  pages={29},
  year={2023},
  publisher={Springer}
}

@article{robotic_perception,
  title={Robotic Perception with a Large Tactile-Vision-Language Model for Physical Property Inference},
  author={Guo, Zexiang and Chen, Hengxiang and Mai, Xinheng and Qiu, Qiusang and Ma, Gan and Kappassov, Zhanat and Li, Qiang and Chen, Nutan},
  journal={arXiv preprint arXiv:2506.19303},
  year={2025}
}

@article{yu2024octopi,
  title={Octopi: Object property reasoning with large tactile-language models},
  author={Yu, Samson and Lin, Kelvin and Xiao, Anxing and Duan, Jiafei and Soh, Harold},
  journal={arXiv preprint arXiv:2405.02794},
  year={2024}
}

@inproceedings{gaussian,
  title={Pugs: Zero-shot physical understanding with gaussian splatting},
  author={Shuai, Yinghao and Yu, Ran and Chen, Yuantao and Jiang, Zijian and Song, Xiaowei and Wang, Nan and Zheng, Jv and Ma, Jianzhu and Yang, Meng and Wang, Zhicheng and others},
  booktitle={2025 IEEE International Conference on Robotics and Automation (ICRA)},
  pages={4478--4485},
  year={2025},
  organization={IEEE}
}

@article{bmw2025predictive,
  title={Predictive visuo-tactile interactive perception framework for object properties inference},
  author={Dutta, Anirvan and Burdet, Etienne and Kaboli, Mohsen},
  journal={IEEE Transactions on Robotics},
  volume={41},
  pages={1386--1403},
  year={2025},
  publisher={IEEE}
}

@inproceedings{wang2020swingbot,
  title={Swingbot: Learning physical features from in-hand tactile exploration for dynamic swing-up manipulation},
  author={Wang, Chen and Wang, Shaoxiong and Romero, Branden and Veiga, Filipe and Adelson, Edward},
  booktitle={2020 IEEE/RSJ International Conference on Intelligent Robots and Systems (IROS)},
  pages={5633--5640},
  year={2020},
  organization={IEEE}
}

@inproceedings{kruzliak2024interactive,
  title={Interactive learning of physical object properties through robot manipulation and database of object measurements},
  author={Kruzliak, Andrej and Hartvich, Jiri and Patni, Shubhan P and Rustler, Lukas and Behrens, Jan Kristof and Abu-Dakka, Fares J and Mikolajczyk, Krystian and Kyrki, Ville and Hoffmann, Matej},
  booktitle={2024 IEEE/RSJ International Conference on Intelligent Robots and Systems (IROS)},
  pages={7596--7603},
  year={2024},
  organization={IEEE}
}

@article{sundaralingam2021hand,
  title={In-hand object-dynamics inference using tactile fingertips},
  author={Sundaralingam, Balakumar and Hermans, Tucker},
  journal={IEEE Transactions on Robotics},
  volume={37},
  number={4},
  pages={1115--1126},
  year={2021},
  publisher={IEEE}
}

@inproceedings{resnet,
  title={Identity mappings in deep residual networks},
  author={He, Kaiming and Zhang, Xiangyu and Ren, Shaoqing and Sun, Jian},
  booktitle={European conference on computer vision},
  pages={630--645},
  year={2016},
  organization={Springer}
}

@inproceedings{bmw2023push,
  title={Push to know!-visuo-tactile based active object parameter inference with dual differentiable filtering},
  author={Dutta, Anirvan and Burdet, Etienne and Kaboli, Mohsen},
  booktitle={2023 IEEE/RSJ International Conference on Intelligent Robots and Systems (IROS)},
  pages={3137--3144},
  year={2023},
  organization={IEEE}
}

@article{xu2019densephysnet,
  title={Densephysnet: Learning dense physical object representations via multi-step dynamic interactions},
  author={Xu, Zhenjia and Wu, Jiajun and Zeng, Andy and Tenenbaum, Joshua B and Song, Shuran},
  journal={arXiv preprint arXiv:1906.03853},
  year={2019}
}

@inproceedings{yang2025learning,
  title={Learning object compliance via young’s modulus from single grasps using camera-based tactile sensors},
  author={Burgess, Michael and Zhao, Jialiang and Willemet, Laurence},
  booktitle={2025 IEEE/RSJ International Conference on Intelligent Robots and Systems (IROS)},
  pages={18535--18542},
  year={2025},
  organization={IEEE}
}

@inproceedings{simreal2025learning,
  title={Learning Object Properties Using Robot Proprioception via Differentiable Robot-Object Interaction},
  author={Chen, Peter Yichen and Liu, Chao and Ma, Pingchuan and Eastman, John and Rus, Daniela and Randle, Dylan and Ivanov, Yuri and Matusik, Wojciech},
  booktitle={2025 IEEE International Conference on Robotics and Automation (ICRA)},
  pages={5997--6004},
  year={2025},
  organization={IEEE}
}

@article{chuo2016learning,
  title={Learning to poke by poking: Experiential learning of intuitive physics},
  author={Agrawal, Pulkit and Nair, Ashvin V and Abbeel, Pieter and Malik, Jitendra and Levine, Sergey},
  journal={Advances in neural information processing systems},
  volume={29},
  year={2016}
}

@article{sim2023object,
  title={Object recognition using mechanical impact, viscoelasticity, and surface friction during interaction},
  author={Uttayopas, Pakorn and Cheng, Xiaoxiao and Eden, Jonathan and Burdet, Etienne},
  journal={IEEE Transactions on Haptics},
  volume={16},
  number={2},
  pages={251--260},
  year={2023},
  publisher={IEEE}
}

@inproceedings{sugaiwa2010methodology,
  title={A methodology for setting grasping force for picking up an object with unknown weight, friction, and stiffness},
  author={Sugaiwa, Taisuke and Fujii, Genki and Iwata, Hiroyasu and Sugano, Shigeki},
  booktitle={2010 10th IEEE-RAS International Conference on Humanoid Robots},
  pages={288--293},
  year={2010},
  organization={IEEE}
}

@inproceedings{guo2025phygrasp,
  title={Phygrasp: generalizing robotic grasping with physics-informed large multimodal models},
  author={Guo, Dingkun and Xiang, Yuqi and Zhao, Shuqi and Zhu, Xinghao and Tomizuka, Masayoshi and Ding, Mingyu and Zhan, Wei},
  booktitle={2025 IEEE/RSJ International Conference on Intelligent Robots and Systems (IROS)},
  pages={14915--14922},
  year={2025},
  organization={IEEE}
}

@article{2025softgrasp,
  title={SoftGrasp: Adaptive grasping for dexterous hand based on multimodal imitation learning},
  author={Li, Yihong and Guo, Ce and Ren, Junkai and Chen, Bailiang and Cheng, Chuang and Zhang, Hui and Lu, Huimin},
  journal={Biomimetic Intelligence and Robotics},
  volume={5},
  number={2},
  pages={100217},
  year={2025},
  publisher={Elsevier}
}

@article{2025fbi,
  title={FBI: Learning Dexterous In-hand Manipulation with Dynamic Visuotactile Shortcut Policy},
  author={Chen, Yijin and Xu, Wenqiang and Yu, Zhenjun and Tang, Tutian and Li, Yutong and Yao, Siqiong and Lu, Cewu},
  journal={arXiv preprint arXiv:2508.14441},
  year={2025}
}

@article{2025omnivtla,
  title={Omnivtla: Vision-tactile-language-action model with semantic-aligned tactile sensing},
  author={Cheng, Zhengxue and Zhang, Yiqian and Zhang, Wenkang and Li, Haoyu and Wang, Keyu and Song, Li and Zhang, Hengdi},
  journal={arXiv preprint arXiv:2508.08706},
  year={2025}
}

@article{2025vitacformer,
  title={ViTacFormer: Learning cross-modal representation for visuo-tactile dexterous manipulation},
  author={Heng, Liang and Geng, Haoran and Zhang, Kaifeng and Abbeel, Pieter and Malik, Jitendra},
  journal={arXiv preprint arXiv:2506.15953},
  year={2025}
}

@article{2505forcevla,
  title={Forcevla: Enhancing vla models with a force-aware moe for contact-rich manipulation (2025)},
  author={Yu, Jiawen and Liu, Hairuo and Yu, Qiaojun and Ren, Jieji and Hao, Ce and Ding, Haitong and Huang, Guangyu and Huang, Guofan and Song, Yan and Cai, Panpan and others},
  journal={arXiv preprint arXiv:2505.22159},
  year={2025}
}

@article{FTACT,
  title={FTACT: Force Torque aware Action Chunking Transformer for Pick-and-Reorient Bottle Task},
  author={Watanabe, Ryo and Alvarez, Maxime and Ferreiro, Pablo and Savkin, Pavel and Sano, Genki},
  journal={arXiv preprint arXiv:2509.23112},
  year={2025}
}

@article{act,
  title={Learning fine-grained bimanual manipulation with low-cost hardware},
  author={Zhao, Tony Z and Kumar, Vikash and Levine, Sergey and Finn, Chelsea},
  journal={arXiv preprint arXiv:2304.13705},
  year={2023}
}

@inproceedings{image2mass,
  title={image2mass: Estimating the Mass of an Object from Its Image},
  author={Standley, Trevor and Sener, Ozan and Chen, Dawn and Savarese, Silvio},
  booktitle={Proceedings of the 1st Annual Conference on Robot Learning},
  pages={324--333},
  year={2017},
  volume={78},
  series={Proceedings of Machine Learning Research},
  publisher={PMLR}
}

@article{egomimic,
  title={EgoMimic: Scaling Imitation Learning via Egocentric Video},
  author={Kareer, Simar and Patel, Dhruv and Punamiya, Ryan and Mathur, Pranay and Cheng, Shuo and Wang, Chen and Hoffman, Judy and Xu, Danfei},
  journal={arXiv preprint arXiv:2410.24221},
  year={2024}
}

@article{egoscale,
  title={EgoScale: Scaling Dexterous Manipulation with Diverse Egocentric Human Data},
  author={Zheng, Ruijie and Niu, Dantong and Xie, Yuqi and Wang, Jing and Xu, Mengda and Jiang, Yunfan and Castaneda, Fernando and Hu, Fengyuan and Tan, You Liang and Fu, Letian and Darrell, Trevor and Huang, Furong and Zhu, Yuke and Xu, Danfei and Fan, Linxi},
  journal={arXiv preprint arXiv:2602.16710},
  year={2026}
}

@article{wang2025tacrefinenet,
  title={TacRefineNet: Tactile-Only Grasp Refinement Between Arbitrary In-Hand Object Poses},
  author={Wang, Shuaijun and Zhou, Haoran and Xiang, Diyun and You, Yangwei},
  journal={arXiv preprint arXiv:2509.25746},
  year={2025}
}

@article{wang2022learning,
  title={Learning adaptive grasping from human demonstrations},
  author={Wang, Shuaijun and Hu, Wenbin and Sun, Lining and Wang, Xin and Li, Zhibin},
  journal={IEEE/ASME Transactions on Mechatronics},
  volume={27},
  number={5},
  pages={3865--3873},
  year={2022},
  publisher={IEEE}
}

\end{document}